\documentclass[pmlr]{jmlr}

\usepackage{longtable}

 \usepackage{booktabs}
 \usepackage{multirow}
 \usepackage{makecell}
 \usepackage{arydshln}
 \usepackage[flushleft]{threeparttable}
 
\usepackage[load-configurations=version-1]{siunitx} 

\makeatletter
\def\set@curr@file#1{\def\@curr@file{#1}} 
\makeatother

\theorembodyfont{\upshape}
\theoremheaderfont{\scshape}
\theorempostheader{:}
\theoremsep{\newline}

\jmlrvolume{219}
\jmlryear{2023}
\jmlrworkshop{Machine Learning for Healthcare}

\title[Large Language Models in Healthcare: A Comparative Study]{Are Large Language Models Ready for Healthcare? A Comparative Study on Clinical Language Understanding}

\author{\Name{Yuqing Wang}
       \Email{wang603@ucsb.edu}\\ 
       \addr Computer Science Department\\
       University of California, Santa Barbara\\
       \AND
       \Name{Yun Zhao}
       \Email{yunzhao20@meta.com}\\ 
       \addr Meta Platforms, Inc.\\
       \AND
       \Name{Linda Petzold}
       \Email{petzold@cs.ucsb.edu}\\ 
       \addr Computer Science Department\\
       University of California, Santa Barbara\\
       }

\begin{document}

\maketitle

\begin{abstract}
Large language models (LLMs) have made significant progress in various domains, including healthcare. However, the specialized nature of clinical language understanding tasks presents unique challenges and limitations that warrant further investigation. In this study, we conduct a comprehensive evaluation of state-of-the-art LLMs, namely GPT-3.5, GPT-4, and Bard, within the realm of clinical language understanding tasks. These tasks span a diverse range, including named entity recognition, relation extraction, natural language inference, semantic textual similarity, document classification, and question-answering. We also introduce a novel prompting strategy, self-questioning prompting (SQP), tailored to enhance the performance of LLMs by eliciting informative questions and answers pertinent to the clinical scenarios at hand. Our evaluation highlights the importance of employing task-specific learning strategies and prompting techniques, such as SQP, to maximize the effectiveness of LLMs in healthcare-related tasks. Our study emphasizes the need for cautious implementation of LLMs in healthcare settings, ensuring a collaborative approach with domain experts and continuous verification by human experts to achieve responsible and effective use, ultimately contributing to improved patient care. Our code is available at \url{https://github.com/EternityYW/LLM_healthcare}.
\end{abstract}

\section{Introduction}
Recent advancements in clinical language understanding hold the potential to revolutionize healthcare by facilitating the development of intelligent systems that support decision-making~\citep{lederman2022tasks, zuheros2021sentiment}, expedite diagnostics~\citep{wang2022exploring, wang2022integrating}, and improve patient care~\citep{christensen2002mplus}. Such systems could assist healthcare professionals in managing the ever-growing body of medical literature, interpreting complex patient records, and developing personalized treatment plans~\citep{pivovarov2015automated, zeng2021natural}. State-of-the-art large language models (LLMs) like OpenAI's GPT-3.5 and GPT-4~\citep{openai2023gpt4}, and Google AI's Bard~\citep{elias2023google}, have gained significant attention for their remarkable performance across diverse natural language understanding tasks, such as sentiment analysis, machine translation, text summarization, and question-answering~\citep{zhong2023can, jiao2023chatgpt, wang2023chatgpt}. However, a comprehensive evaluation of their effectiveness in the specialized healthcare domain, with its unique challenges and complexities, remains necessary.

The healthcare domain presents distinct challenges, including handling specialized medical terminology, managing the ambiguity and variability of clinical language, and meeting the high demands for reliability and accuracy in critical tasks. Although existing research has explored the application of LLMs in healthcare, the focus has typically been on a limited set of tasks or learning strategies. For example, studies have investigated tasks like medical concept extraction, patient cohort identification, and drug-drug interaction prediction, primarily relying on supervised learning approaches~\citep{vilar2018detection, gehrmann2018comparing, afshar2019natural}. In this study, we broaden this scope by evaluating LLMs on various clinical language understanding tasks, including natural language inference (NLI), document classification, semantic textual similarity (STS), question-answering (QA), named entity recognition (NER), and relation extraction.

Furthermore, the exploration of learning strategies such as few-shot learning, transfer learning, and unsupervised learning in the healthcare domain has been relatively limited. Similarly, the impact of diverse prompting techniques on improving model performance in clinical tasks has not been extensively examined, leaving room for a comprehensive comparative study.

In this study, we aim to bridge this gap by evaluating the performance of state-of-the-art LLMs on a range of clinical language understanding tasks. LLMs offer the exciting prospect of in-context few-shot learning via prompting, enabling task completion without fine-tuning separate language model checkpoints for each new challenge. In this context, we propose a novel prompting strategy called self-questioning prompting (SQP) to enhance these models' effectiveness across various tasks. Our empirical evaluations demonstrate the potential of SQP as a promising technique for improving LLMs in the healthcare domain. Furthermore, by pinpointing tasks where the models excel and those where they struggle, we highlight the need for addressing specific challenges such as wording ambiguity, lack of context, and negation handling, while emphasizing the importance of responsible LLM implementation and collaboration with domain experts in healthcare settings.

In summary, our contributions are threefold:
\begin{enumerate}
\item[(1)] To the best of our knowledge, this is the first comparative study to investigate the effectiveness of state-of-the-art LLMs on a variety of clinical language understanding tasks with diverse learning strategies and prompting strategies.
\item[(2)] We introduce a novel prompting strategy, namely self-questioning prompting, which aims to enhance the performance of LLMs by encouraging the generation of informative questions and answers and prompting a deeper understanding of the medical scenarios being described.
\item[(3)] Our error analysis on the most challenging task common to all models highlights the unique challenges each model faces, including wording ambiguity, lack of context, and negation, emphasizing the need for a cautious approach when employing LLMs in healthcare as a supplement to human expertise.
\end{enumerate}

\subsection*{Generalizable Insights about Machine Learning in the Context of Healthcare}
Our study presents a comprehensive evaluation of state-of-the-art LLMs in the healthcare domain, examining their capabilities and limitations across a variety of clinical language understanding tasks. We develop and demonstrate the efficacy of our self-questioning prompting (SQP) strategy, which involves generating context-specific questions and answers to guide the model towards a better understanding of clinical scenarios. This tailored learning approach significantly enhances LLM performance in healthcare-focused tasks. Our in-depth error analysis on the most challenging task shared by all models uncovers unique difficulties encountered by each model, such as wording ambiguity, lack of context, and negation issues. These findings emphasize the need for a cautious approach when implementing LLMs in healthcare as a complement to human expertise. We underscore the importance of integrating domain-specific knowledge, fostering collaborations among researchers, practitioners, and domain experts, and employing task-oriented prompting techniques like SQP. By addressing these challenges and harnessing the potential benefits of LLMs, we can contribute to improved patient care and clinical decision-making in healthcare settings.

\section{Related Work}
In this section, we review the relevant literature on large language models applied to clinical language understanding tasks in healthcare, as well as existing prompting strategies.

\subsection{Large Language Models in Healthcare}
The advent of the Transformer architecture~\citep{vaswani2017attention} revolutionized the field of natural language processing, paving the way for the development of large-scale pre-trained language models such as base BERT~\citep{devlin2018bert} and RoBERTa~\citep{liu2019roberta}. In the healthcare domain, domain-specific adaptations of BERT, such as BioBERT~\citep{lee2020biobert} and ClinicalBERT~\citep{alsentzer2019publicly}, have been introduced to tackle various clinical language understanding tasks. More recently, GPT-3.5 and its successor GPT-4, launched by OpenAI~\citep{openai2023gpt4}, as well as Bard, developed by Google AI~\citep{elias2023google}, have emerged as state-of-the-art LLMs, showcasing impressive capabilities in a wide range of applications, including healthcare~\citep{biswas2023chatgpt, kung2023performance, patel2023chatgpt, singhal2023large}.

Clinical language understanding is a critical aspect of healthcare informatics, focused on extracting meaningful information from diverse sources, such as electronic health records~\citep{juhn2020artificial}, scientific articles~\citep{grabar2021year}, and patient-authored text data~\citep{mukhiya2020adaptation}. This domain encompasses various tasks, including NER~\citep{nayel2017improving}, relation extraction~\citep{lv2016clinical}, NLI~\citep{romanov2018lessons}, STS~\citep{wang2020learning}, document classification~\citep{hassanzadeh2018transferability}, and QA~\citep{soni2020evaluation}. Prior work has demonstrated the effectiveness of domain-specific models in achieving improved performance on these tasks compared to general-purpose counterparts~\citep{peng2019transfer, mascio2020comparative, digan2021can}. However, challenges posed by complex medical terminologies, the need for precise inference, and the reliance on domain-specific knowledge can limit their effectiveness~\citep{shen2023chatgpt}. In this work, we address some of these limitations by conducting a comprehensive evaluation of state-of-the-art LLMs  on a diverse set of clinical language understanding tasks, focusing on their performance and applicability within healthcare settings.

\subsection{Prompting Strategies}
Prompting strategies, often used in conjunction with few-shot or zero-shot learning~\citep{brown2020language, kojima2022large}, guide and refine the behavior of LLMs to improve performance on various tasks. In these learning paradigms, LLMs are conditioned on a limited number of examples in the form of prompts, enabling them to generalize and perform well on the target task. Standard prompting techniques~\citep{brown2020language} involve providing an LLM with a clear and concise prompt, often in the form of a question or statement, which directs the model towards the desired output. Another approach, known as chain-of-thought prompting~\citep{weichain, kojima2022large}, leverages a series of interconnected prompts to generate complex reasoning or multi-step outputs. While these existing prompting strategies have shown considerable success, their effectiveness can be limited by the quality and informativeness of the prompts~\citep{wang2022iteratively}, which may not always capture the intricate nuances of specialized domains like healthcare. Motivated by these limitations, we propose a novel prompting strategy called self-questioning prompting (SQP). SQP aims to enhance the performance of LLMs by generating informative questions and answers related to the given clinical scenarios, thus addressing the unique challenges of the healthcare domain and contributing to improved task-specific performance.

\section{Self-Questioning Prompting}
\begin{figure*}[htbp]
\centering
\includegraphics[width=0.75\textwidth]{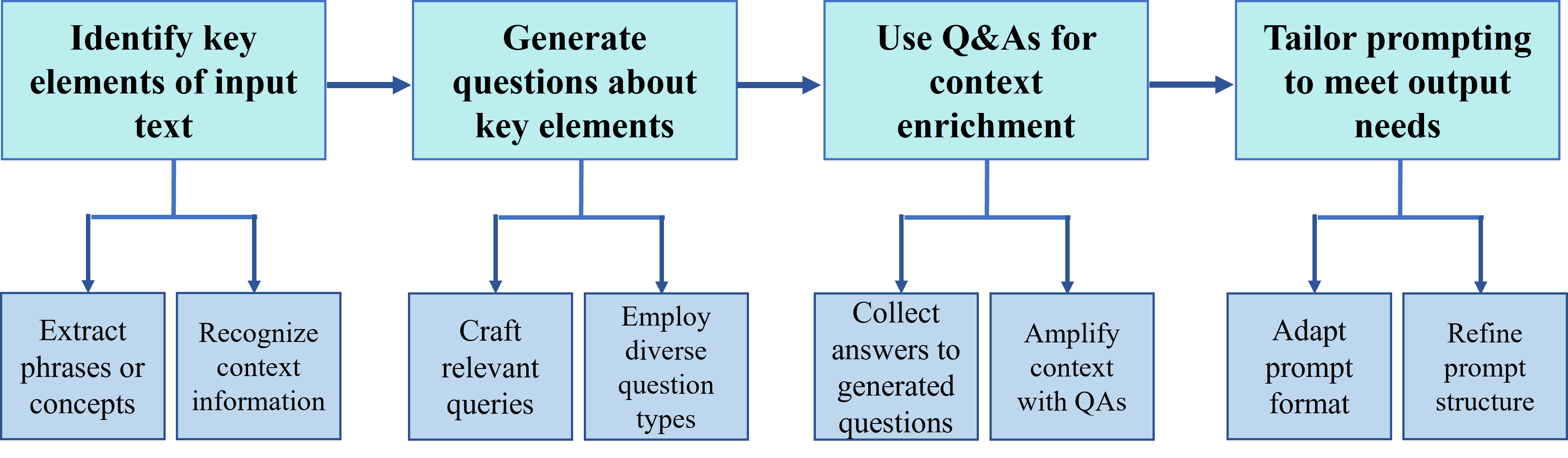}  
\caption{Construction process of self-questioning prompting (SQP).}
\label{fig: SQP_construction}
\end{figure*}

Complex problems can be daunting, but they can often be solved by breaking them down into smaller parts and asking questions to clarify understanding and explore different aspects. Inspired by this human-like reasoning process, we introduce a novel method called self-questioning prompting (SQP) for LLMs. SQP aims to enhance model performance by encouraging models to be more aware of their own thinking processes, enabling them to better understand relevant concepts and develop deeper comprehension. This is achieved through the generation of targeted questions and answers that provide additional context and clarification, ultimately leading to improved performance on various tasks. The general construction process of SQP for a task, as shown in Figure~\ref{fig: SQP_construction}, involves identifying key information in the input text, generating targeted questions to clarify understanding, using the questions and answers to enrich the context of the task prompt, and tailoring the strategy to meet the unique output requirements of each task.
For a better understanding of the general construction procedure, consider an example prompting for NLI task:
\begin{enumerate}
    \item[1.] \textbf{Key Information:} With two clinical sentences, \{sentence\_1\} and \{sentence\_2\}, the model is asked to ``Generate questions about the medical situations described". This prompt guides the model to identify important elements.
    \item[2.] \textbf{Question Generation:} Following the first prompt, the model creates questions about the identified details, solidifying its grasp on the context.
    \item[3.] \textbf{Enriching Context:} The model then ``Answer these questions using basic medical knowledge and use the insights to evaluate their relationship". This prompt instructs the model to deepen its understanding.
    \item[4.] \textbf{Task-Specific Strategy:} Lastly, the model follows the prompt to ``Categorize the relationship between \{sentence\_1\} and \{sentence\_2\} as entailment if \{sentence\_2\} logically follows \{sentence\_1\}, contradiction if they oppose each other, or neutrality if unrelated". This directly links the task requirements with the model's understanding.
\end{enumerate}

\begin{figure*}[htbp]
\centering
\includegraphics[width=\textwidth]{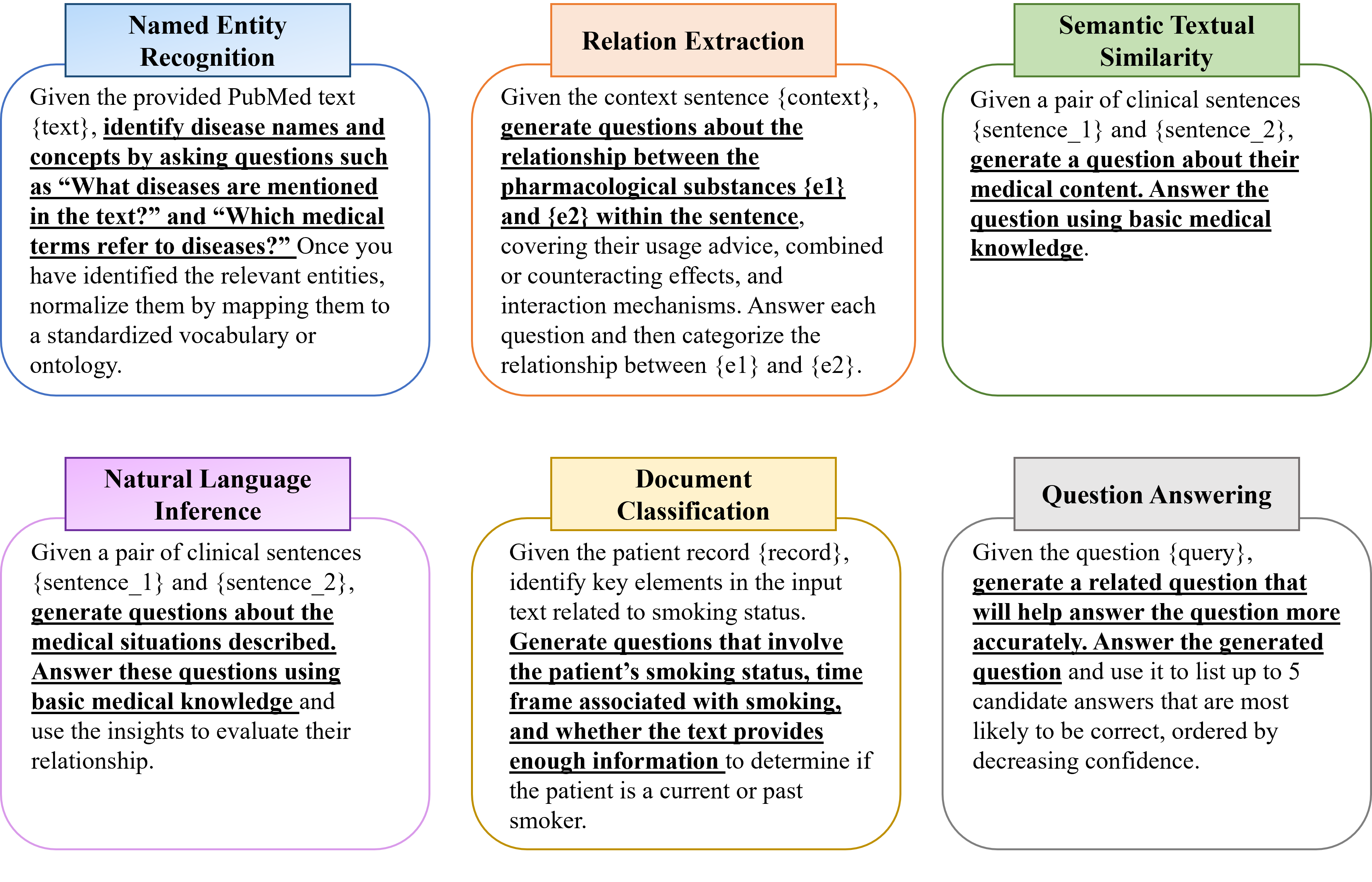}  
\caption{Self-questioning prompting (SQP) templates for six clinical language understanding tasks, with the core self-questioning process underscored and bolded. These components represent the generation of targeted questions and answers, guiding the model's reasoning and enhancing task performance.}
\label{fig: SQP_template}
\end{figure*}

In Table~\ref{tab: comparison_prompting}, we compare the proposed SQP with existing prompting methods, including standard prompting and chain-of-thought prompting, highlighting the differences in guidelines and purposes for each strategy. Subsequently, we present the SQP templates for six clinical language understanding tasks. The core self-questioning process is highlighted in each template, as shown in Figure~\ref{fig: SQP_template}. The SQP templates were developed through a combination of consultations with healthcare professionals and iterative testing. We evaluated multiple prompt candidates, with the best-performing templates chosen for use in the study. In the case of few-shot examples, the SQP QA pairs were annotated by healthcare professionals for model input. These underscored and bold parts illustrate how SQP generates targeted questions and answers related to the tasks, which guide the model's reasoning, leading to improved task performance. By incorporating this self-questioning process into the prompts, SQP enables the model to utilize its knowledge more effectively and adapt to a wide range of clinical tasks.

\begin{table*}[htbp]
\centering 
  \caption{Comparison among standard prompting, chain-of-thought prompting, and self-questioning prompting.}
  \begin{tabular}{ccc}
    \toprule
 \textbf{Prompting Strategy} & \textbf{Guideline} & \textbf{Purpose} \\ \midrule Standard & \makecell{Use a direct, concise \\ prompt for the desired task.} & \makecell{To obtain a direct \\ response from the model.} \\  \midrule Chain-of-Thought & \makecell{Create interconnected \\ prompts guiding the model \\ through logical reasoning.} & \makecell{To engage the model's \\ reasoning by breaking \\ down complex tasks.} \\ \midrule Self-Questioning & \makecell{Generate targeted questions \\ and use answers to guide \\ the task response.} & \makecell{To deepen the model's \\ understanding and \\ enhance performance.}\\
  \bottomrule
\end{tabular}
\label{tab: comparison_prompting}
\end{table*}

\section{Datasets}
We utilize a wide range of biomedical and clinical language understanding datasets for our experiments. These datasets encompass various tasks, including NER (NCBI-Disease~\citep{dougan2014ncbi} and BC5CDR-Chem~\citep{li2016biocreative}), relation extraction (i2b2 2010-Relation~\citep{uzuner20112010} and SemEval 2013-DDI~\citep{segura2013semeval}), STS (BIOSSES~\citep{souganciouglu2017biosses}), NLI (MedNLI~\citep{romanov2018lessons}), document classification (i2b2 2006-Smoking~\citep{uzuner2006i2b2}), and QA (bioASQ 10b-Factoid~\citep{tsatsaronis2015overview}). Among these tasks, STS (BIOSSES) is a regression task, while the rest are classification tasks. Table~\ref{tab: dataset} offers a comprehensive overview of the tasks and datasets. For NER tasks, we adopt the BIO tagging scheme, where ‘B' represents the beginning of an entity, ‘I' signifies the continuation of an entity, and ‘O' denotes the absence of an entity. The output column in Table~\ref{tab: dataset} presents specific classes, scores, or tagging schemes associated with each task.

For relation extraction, SemEval 2013-DDI requires identifying one of the following labels: Advice, Effect, Mechanism, or Int. In the case of i2b2 2010-Relation, it necessitates predicting relationships such as Treatment Improves Medical Problem (TrIP), Treatment Worsens Medical Problem (TrWP), Treatment Causes Medical Problem (TrCP), Treatment is Administered for Medical Problem (TrAP), Treatment is Not Administered because of Medical Problem (TrNAP), Test Reveals Medical Problem (TeRP), Test Conducted to Investigate Medical Problem (TeCP), or Medical Problem Indicates Medical Problem (PIP).

\begin{table*}[htbp]
\centering 
  \caption{Overview of biomedical/clinical language understanding tasks and datasets.}
  \resizebox{\linewidth}{!}{%
  \begin{tabular}{cccc}
    \toprule
  \textbf{Task} & \textbf{Dataset} & \textbf{Output} & \textbf{Metric} \\ \midrule
  \makecell{Named Entity \\ Recognition} & \makecell{NCBI-Disease, \\ BC5CDR-Chemical} & \makecell{BIO tagging for \\diseases and chemicals} & Micro F1 \\ \midrule Relation Extraction & \makecell {i2b2 2010-Relation, \\ SemEval 2013-DDI} & \makecell{relations between entities} & \makecell{Micro F1, \\ Macro F1} \\ \midrule \makecell{Semantic Textual \\ Similarity} & BIOSSES & \makecell{similarity scores from 0 \\ (different) to 4 (identical)} & \makecell{Pearson \\ Correlation} \\ \midrule \makecell{Natural Language \\ Inference} & MedNLI & \makecell{entailment, neutral, \\ contradiction} & Accuracy \\ \midrule \makecell{Document \\ Classification} & i2b2 2006-Smoking  & \makecell{current smoker, past smoker, \\ smoker, non-smoker, unknown} & Micro F1 \\ \midrule Question-Answering & bioASQ 10b-Factoid & factoid answers & \makecell{ Mean Reciprocal Rank, \\ Lenient Accuracy} \\

  \bottomrule
\end{tabular}}

\label{tab: dataset}
\end{table*}

\section{Experiments}
In this section, we outline the experimental setup and evaluation procedure used to evaluate the performance of various LLMs on tasks related to biomedical and clinical text comprehension and analysis.

\subsection{Experimental Setup}
We investigate various prompting strategies for state-of-the-art LLMs, employing N-shot learning techniques on diverse clinical language understanding tasks.

\indent \textbf{Large Language Models.}
We assess the performance of three state-of-the-art LLMs, each offering unique capabilities and strengths. First, we examine GPT-3.5, an advanced model developed by OpenAI, known for its remarkable language understanding and generation capabilities. Next, we investigate GPT-4, an even more powerful successor to GPT-3.5, designed to push the boundaries of natural language processing further. Finally, we explore Bard, an innovative language model launched by Google AI. We experiment with these models through their web versions. By comparing these models, we aim to gain insights into their performance on clinical language understanding tasks.

\indent \textbf{Prompting Strategies.}
We employ three prompting strategies to optimize the performance of LLMs on each task: standard prompting, chain-of-thought prompting, and our proposed self-questioning prompting. Standard prompting serves as the baseline, while chain-of-thought and self-questioning prompting techniques are investigated to assess their potential impact on model performance. The full set of prompting templates used for each task are given in Appendix~\ref{prompt_template}.

\indent \textbf{N-Shot Learning.}
We explore N-shot learning for LLMs, focusing on zero-shot and 5-shot learning scenarios. Zero-shot learning refers to the situation where the model has not been exposed to any labeled examples during training and is expected to generalize to the task without prior knowledge. In contrast, 5-shot learning involves the model receiving a small amount of labeled data, consisting of five few-shot exemplars from the training set, to facilitate its adaptation to the task. We evaluate the model's performance in both zero-shot and 5-shot learning settings to understand its ability to generalize and adapt to different tasks in biomedical and clinical domains.

\subsection{Evaluation Procedure}
To assess the performance for each task, given the constraints of model release timings and web version utilization, we form an evaluation set by randomly selecting 50\% of instances from the original test set. In the case of zero-shot learning, we directly evaluate the model's performance on this evaluation set. For 5-shot learning, we enhance the model with five few-shot exemplars, which are randomly chosen from the training set. The model's performance is then assessed using the same evaluation set as in the zero-shot learning scenario.

\section{Results}
In this section, we present a comprehensive analysis of the performance of the LLMs (i.e., Bard, GPT-3.5, and GPT-4) on clinical language understanding tasks. We begin by comparing the overall performance of these models, followed by an examination of the effectiveness of various prompting strategies. Next, we delve into a detailed task-by-task analysis, providing insights into the models' strengths and weaknesses across different tasks. Finally, we conduct a case study on error analysis, investigating common error types and the potential improvements brought about by advanced prompting techniques.

\subsection{Overall Performance Comparison}
In our study, we evaluate the performance of Bard, GPT-3.5, and GPT-4 on various clinical benchmark datasets spanning multiple tasks. We employ different prompting strategies, including standard, chain-of-thought, and self-questioning, as well as N-shot learning with N equal to 0 and 5. Table~\ref{tab: overall_performance} summarizes the experimental results.

We observe that GPT-4 generally outperforms Bard and GPT-3.5 in tasks involving the identification and classification of specific information within text, such as NLI (MedNLI), NER (NCBI-Disease, BC5CDR-Chemical), and STS (BIOSSES). In the realm of document classification, a task that involves assigning predefined categories to entire documents, GPT-4 also surpasses GPT-3.5 and Bard on the i2b2 2006-Smoking dataset. In relation extraction, GPT-4 outperforms both Bard and GPT-3.5 on the SemEval 2013-DDI dataset, while Bard demonstrates superior performance in the i2b2 2010-Relation dataset. Additionally, Bard excels in tasks that require a more factual understanding of the text, such as QA (BioASQ 10b-Factoid).

Regarding prompting strategies, self-questioning consistently outperforms standard prompting and exhibits competitive performance when compared to chain-of-thought prompting across all settings. Our findings suggest that self-questioning is a promising approach for enhancing the performance of LLMs, achieving the best performance for the majority of tasks, except for QA (BioASQ 10b-Factoid).

Furthermore, our study demonstrates that 5-shot learning generally leads to improved performance across all tasks when compared to zero-shot learning, although not universally. This finding indicates that incorporating even a modest amount of task-specific training data can substantially enhance the effectiveness of pre-trained LLMs.

\begin{table*}[tp]
\centering 
  \caption{Performance comparison of Bard, GPT-3.5, and GPT-4 with different prompting strategies (standard, chain-of-thought, and self-questioning) and N-shot learning (N = 0, 5) on clinical benchmark datasets. randomly sampled evaluation data from the test set. Our results show that GPT-4 outperforms Bard and GPT-3.5 in tasks that involve identification and classification of specific information within text, while Bard achieves higher accuracy than GPT-3.5 and GPT-4 on tasks that require a more factual understanding of the text. Additionally, self-questioning prompting consistently achieves the best performance on the majority of tasks. The best results for each dataset are highlighted in bold.}
  \resizebox{\linewidth}{!}{%
  \begin{tabular}{cccccccccc}

    \toprule
    \textbf{Model} & \makecell{\textbf{NCBI-} \\ \textbf{Disease}}& \makecell{\textbf{BC5CDR-} \\ \textbf{Chemical}} & \makecell{\textbf{i2b2 2010-} \\ \textbf{Relation}} & \makecell{\textbf{SemEval 2013-} \\ \textbf{DDI}} & \textbf{BIOSSES} & \textbf{MedNLI} & \makecell{\textbf{i2b2 2006-} \\ \textbf{Smoking}} & \multicolumn{2}{c}{\makecell{\textbf{BioASQ 10b-} \\ \textbf{Factoid}}} \\  & \textit{Micro F1} & \textit{Micro F1} & \textit{Micro F1} & \textit{Macro F1} & \textit{Pear.} & \textit{Acc.} & \textit{Micro F1} & \textit{MRR} & \textit{Len. Acc.} \\
    \midrule
    Bard & & & & & & & & & \\
    \hdashline
    w/ zero-shot StP & 0.911 & 0.947 & 0.720 & 0.490 & 0.401 & 0.580 & 0.780 & 0.800 & 0.820 \\
    w/ 5-shot StP & 0.933 & 0.972 & 0.900 & 0.528 & 0.449 & 0.640 & 0.820 & 0.845 & 0.880 \\
    w/ zero-shot CoTP & 0.946 & 0.972 & 0.660 & 0.525 & 0.565 & 0.580 & 0.760 & \textbf{0.887} & \textbf{0.920} \\
    w/ 5-shot CoTP & 0.955 & 0.977 & 0.900 & 0.709 & 0.602 & 0.720 & 0.800 & 0.880 & 0.900\\
    w/ zero-shot SQP & 0.956 & 0.977 & 0.760 & 0.566 & 0.576 & 0.760 & 0.760 & 0.850 & 0.860 \\
    w/ 5-shot SQP & 0.960 & 0.983 & \textbf{0.940} & 0.772 & 0.601 & 0.760 & 0.820 & 0.860 & 0.860\\
    \midrule
    GPT-3.5 & & & & & & & & & \\
    \hdashline
    w/ zero-shot StP & 0.918 & 0.939 & 0.780 & 0.360 & 0.805 & 0.700 & 0.680 & 0.707 & 0.720 \\
    w/ 5-shot StP & 0.947 & 0.967 & 0.840 & 0.531 & 0.828 & 0.780 & 0.780 & 0.710 & 0.740\\
    w/ zero-shot CoTP & 0.955 & 0.977 & 0.680 & 0.404 & 0.875 & 0.740 & 0.680 & 0.743 & 0.800\\
    w/ 5-shot CoTP & 0.967 & 0.977 & 0.840 & 0.548 & 0.873 & 0.740 & 0.740 & 0.761 & 0.820\\
    w/ zero-shot SQP & 0.963 & 0.974 & 0.860 & 0.529 & 0.873 & 0.760 & 0.720 & 0.720 & 0.740\\
    w/ 5-shot SQP & 0.970 & 0.983 & 0.860 & 0.620 & 0.892 & 0.820 & 0.820 & 0.747 & 0.780\\
    \midrule
    GPT-4 & & & & & & & & & \\
    \hdashline
    w/ zero-shot StP & 0.968 & 0.976 & 0.860 & 0.428 & 0.820 & 0.800 & \textbf{0.900} & 0.795 & 0.820\\
    w/ 5-shot StP & 0.975 & 0.989 & 0.860 & 0.502 & 0.848 & 0.840 & 0.880 & 0.815 & 0.840\\
    w/ zero-shot CoTP & 0.981 & 0.994 & 0.860 & 0.509 & 0.875 & 0.840 & 0.860 & 0.805 & 0.840\\
    w/ 5-shot CoTP & 0.984 & 0.994 & 0.880 & 0.544 & 0.897 & 0.800 & 0.860 & 0.852 & 0.880\\
    w/ zero-shot SQP & \textbf{0.985} & 0.992 & 0.920 & 0.595 & 0.889 & \textbf{0.860} & \textbf{0.900} & 0.844 & 0.900 \\
    w/ 5-shot SQP & 0.984 & \textbf{0.995} & 0.920 & \textbf{0.798} & \textbf{0.916} & \textbf{0.860} & 0.860 & 0.873 & 0.900\\
  \bottomrule
\end{tabular}}
\begin{tablenotes}
      \small
      \item \textit{Note:} Acc. = Accuracy; CoTP = Chain-of-Thought Prompting; Len. Acc. = Lenient Accuracy; MRR = Mean Reciprocal Rank; Pear. = Pearson Correlation; StP = Standard Prompting.
    \end{tablenotes}

\label{tab: overall_performance}
\end{table*}

\subsection{Prompting Strategies Comparison}
\begin{figure*}[htbp]
\centering
\includegraphics[width=\textwidth]{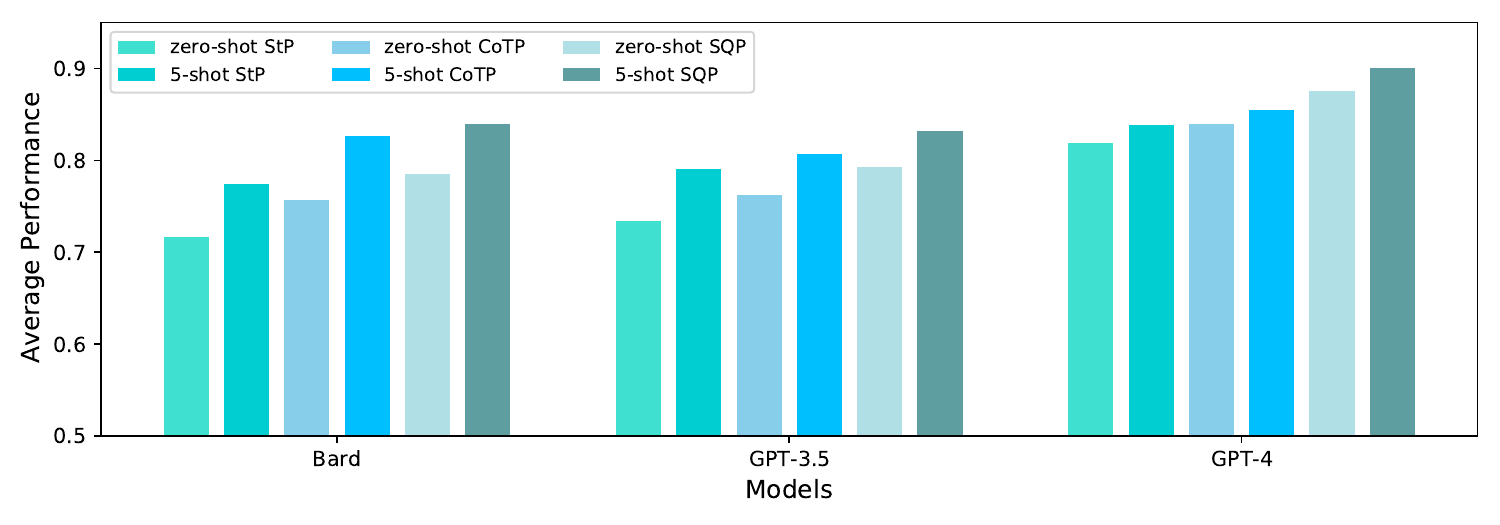}  
\caption{Average performance comparison of three prompting methods in zero-shot and 5-shot learning settings across Bard, GPT-3.5, and GPT-4 models. Performance values are averaged across all datasets, assuming equal importance for datasets and evaluation metrics, as well as direct comparability. The self-questioning prompting method consistently outperforms standard and chain-of-thought prompting, and GPT-4 excels among the models.}
\label{fig: average_performance}
\end{figure*}

We evaluate the performance of different prompting strategies, specifically standard prompting, self-questioning prompting (SQP), and chain-of-thought prompting (CoTP) on both zero-shot and 5-shot learning settings across various models and datasets. Figure~\ref{fig: average_performance} presents the averaged performance comparison over all datasets, under the assumption that datasets and evaluation metrics are equally important and directly comparable. We observe that self-questioning prompting consistently yields the best performance compared to standard and chain-of-thought prompting. In addition, GPT-4 excels among the models, demonstrating the highest overall performance. 

Table~\ref{tab: performance_improvement_0shot} and Table~\ref{tab: performance_improvement_5shot} demonstrate performance improvements of prompting strategies over multiple datasets and models under zero-shot and 5-shot settings, respectively, using standard prompting as a baseline. In the zero-shot learning setting (Table~\ref{tab: performance_improvement_0shot}), self-questioning prompting achieves the highest improvement in the majority of tasks, with improvements ranging from 4.9\% to 46.9\% across different datasets.

In the 5-shot learning setting (Table~\ref{tab: performance_improvement_5shot}), self-questioning prompting leads to the highest improvement in most tasks, with improvements ranging from 2.9\% to 59.0\%. In both settings, we also observe some instances where chain-of-thought or self-questioning prompting yields negative values, such as relation extraction (i2b2 2010-Relation) and document classification (i2b2 2006-Smoking), indicating inferior performance compared to standard prompting. This could be due to the specific nature of certain tasks, where the additional context or complexity introduced by the alternative prompting strategies might not contribute to better understanding or performance. It might also be possible that the model's capacity is insufficient to take advantage of the additional information provided by the alternative prompting strategies in some cases.

Overall, self-questioning prompting generally outperforms other prompting strategies across different models and datasets in both zero-shot and 5-shot learning settings, despite occasional inferior performance in specific tasks. This suggests that self-questioning prompting can be a promising technique for improving performance in the domain of clinical language understanding. Furthermore, GPT-4 emerges as the top-performing model, emphasizing the potential for various applications in the clinical domain.

\begin{table*}[htbp]
\centering 
  \caption{Comparison of zero-shot learning performance improvements (in \%) for different models and prompting techniques on multiple datasets, with standard prompting as the baseline. Bold values indicate the highest improvement for each dataset across models and prompting strategies, while negative values signify inferior performance. Self-questioning prompting leads to the largest improvement in the majority of tasks.}
  \begin{tabular}{cccccccc}
    \toprule
    \multirow{2}{*}{\textbf{Dataset}} & \multirow{2}{*}{\textbf{Metric}} & \multicolumn{2}{c}{\textbf{Bard}} & \multicolumn{2}{c}{\textbf{GPT-3.5}} & \multicolumn{2}{c}{\textbf{GPT-4}} \\
\cmidrule(lr){3-4}\cmidrule(lr){5-6}\cmidrule(lr){7-8}  & & CoTP & SQP & CoTP & SQP & CoTP & SQP  \\ \midrule  NCBI-Disease & \textit{Micro F1} & 3.8 & \textbf{4.9} & 4.0 & \textbf{4.9} & 1.3 & 1.8 \\ BC5CDR-Chemical & \textit{Micro F1} & 2.6 & 3.2 & \textbf{4.0} & 3.7 & 1.8 & 1.6 \\ i2b2 2010-Relation & \textit{Micro F1} & $-8.3$ & 5.6 & $-12.8$ & \textbf{10.3} & 0.0 & 7.0 \\ SemEval 2013-DDI & \textit{Macro F1} & 7.1 & 15.5 & 12.2 & \textbf{46.9} & 18.9 & 39.0 \\ BIOSSES & \textit{Pear.} & 40.9 & \textbf{43.6} & 8.7 & 8.4 & 6.7 & 8.4 \\ MedNLI & \textit{Acc.} & 0.0 & \textbf{31.0} & 5.7 & 8.6 & 5.0 & 7.5 \\ i2b2 2006-Smoking & Micro F1 & $-2.6$ & $-2.6$ & 0.0 & \textbf{5.9} & $-4.4$ & 0.0 \\ BioASQ 10b-Factoid & \textit{MRR} & \textbf{10.9} & 6.3 & 5.1 & 1.8 & 1.3 & 6.2 \\ BioASQ 10b-Factoid & \textit{Len. Acc.} & \textbf{12.2} & 4.9 & 11.1 & 2.8 & 2.4 & 9.8 \\

  \bottomrule
\end{tabular}
\label{tab: performance_improvement_0shot}
\end{table*}

\begin{table*}[htbp]
\centering 
  \caption{Comparison of 5-shot learning performance improvements (in \%) for different models and prompting techniques on multiple datasets, with standard prompting as the baseline. Bold values indicate the highest improvement for each dataset across models and prompting strategies, while negative values signify inferior performance. Self-questioning prompting leads to the highest improvement in 6 out of 8 tasks, followed by chain-of-thought prompting with 2 largest improvements.}
  \begin{tabular}{cccccccc}
    \toprule
    \multirow{2}{*}{\textbf{Dataset}} & \multirow{2}{*}{\textbf{Metric}} & \multicolumn{2}{c}{\textbf{Bard}} & \multicolumn{2}{c}{\textbf{GPT-3.5}} & \multicolumn{2}{c}{\textbf{GPT-4}} \\
\cmidrule(lr){3-4}\cmidrule(lr){5-6}\cmidrule(lr){7-8}  & & CoTP & SQP & CoTP & SQP & CoTP & SQP  \\ \midrule  NCBI-Disease & \textit{Micro F1} & 2.4 & \textbf{2.9} & 2.1 & 2.4 & 0.9 & 0.9 \\ BC5CDR-Chemical & \textit{Micro F1} & 0.5 & 1.1 & 1.0 & \textbf{1.7} & 0.5 & 0.6 \\ i2b2 2010-Relation & \textit{Micro F1} & 0.0 & 4.4 & 0.0 & 2.4 & 2.3 & \textbf{7.0} \\ SemEval 2013-DDI & \textit{Macro F1} & 34.3 & 46.2 & 3.2 & 16.8 & 8.4 & \textbf{59.0} \\ BIOSSES & \textit{Pear.} & \textbf{34.1} & 33.9 & 5.4 & 7.7 & 5.8 & 8.0 \\ MedNLI & \textit{Acc.} & 12.5 & \textbf{18.8} & $-5.1$ & 5.1 & $-4.8$ & 2.4\\ i2b2 2006-Smoking & Micro F1 & $-2.4$ & 0.0 & $-5.1$ & \textbf{5.1} & $-2.3$ & $-2.3$\\ BioASQ 10b-Factoid & \textit{MRR} & 4.1 & 1.8 & \textbf{7.2} & 5.2 & 4.5 & 7.1  \\ BioASQ 10b-Factoid & \textit{Len. Acc.} & 2.3 & $-2.3$ & \textbf{10.8} & 5.4 & 4.8 & 7.1 \\

  \bottomrule
\end{tabular}
\label{tab: performance_improvement_5shot}
\end{table*}

\subsection{Task-by-Task Analysis}
To delve deeper into the specific characteristics and challenges associated with each task (i.e., NER, relation extraction, STS, NLI, document classification, and QA), we individually analyze the results, aiming to better understand the underlying factors that contribute to model performance and identify areas for potential improvement or further investigation. 

\indent \textbf{Named Entity Recognition Task.}
In the NER task, we focus on two datasets: NCBI-Disease and BC5CDR-Chemical. Employing the BIO tagging scheme, we evaluate model performance using the micro F1 metric. NER tasks in the biomedical domain pose unique challenges due to specialized terminology, complex entity names, and frequent use of abbreviations. Our results indicate that, compared to standard prompting, self-questioning prompting leads to average improvements of 3.9\% and 2.8\% in zero-shot learning for NCBI-Disease and BC5CDR-Chemical, respectively. In the 5-shot setting, the average improvements are 2.1\% and 1.1\%, respectively. Moreover, GPT-4 demonstrates the most significant performance boost compared to Bard and GPT-3.5.

We also conduct a qualitative analysis by examining specific examples from the datasets, such as the term ``aromatic ring" in the BC5CDR-Chemical dataset, which is often incorrectly predicted as ``B-Chemical" (beginning of a chemical entity) instead of ``O" (outside of any entity) by the models. This error might occur because the term ``aromatic ring" refers to a structural feature commonly found in chemical compounds, leading models to associate it with chemical entities and misclassify it. This example highlights the challenges faced by the models in accurately recognizing entities, particularly when dealing with terms that have strong associations with specific entity types. It also demonstrates the potential limitations of prompting strategies in addressing these challenges, as models may still struggle to disambiguate such terms, despite employing different prompting techniques.

\indent \textbf{Relation Extraction Task.}
In the relation extraction task involving the i2b2 2010-Relation and SemEval 2013-DDI datasets, we evaluate our model's performance using micro F1 and macro F1 scores, respectively. Our study reveals that self-questioning prompting leads to average improvements of 7.6\% and 33.8\% in zero-shot learning for the i2b2 2010-Relation and SemEval 2013-DDI datasets, respectively. In the 5-shot setting, the average improvements are 4.6\% and 40.7\%, respectively. GPT-4 demonstrates more significant performance improvement compared to Bard and GPT-3.5.

For our qualitative analysis, we examine a challenging example from the i2b2 2010-Relation dataset, where the models struggle to identify the correct relationship between ``Elavil" and ``stabbing left-sided chest pain". The gold label indicates ``TrWP" (Treatment Worsens Medical Problem), but all models incorrectly predict it as ``TrAP" (Treatment is Administered for Medical Problem). This misclassification may arise from the models' inability to recognize that the patient still experiences severe pain despite taking Elavil. This example highlights the difficulties encountered by the models in accurately identifying nuanced relationships in complex biomedical texts. Incorporating domain-specific knowledge could help to better capture the subtleties of such relationships.

\indent \textbf{Semantic Textual Similarity Task.} In the STS task, we focus on the BIOSSES dataset and evaluate our model's performance using Pearson correlation. Our study reveals that self-questioning prompting leads to average improvements of 20.1\% and 16.5\% in zero-shot and 5-shot settings, respectively. GPT-4 outperforms Bard and GPT-3.5 across all settings.

Taking a closer look, we examine a pair of sentences with a gold label similarity score of 0.2, indicating high dissimilarity. The first sentence discusses the specific effect of mutant K-Ras on tumor progression, while the second sentence refers to an important advance in lung cancer research without mentioning any specific details. However, the average score predicted by models, regardless of the setting, is 2.0. This discrepancy may arise from the models' difficulty in grasping the distinct contexts in which the sentences are written. The models might be misled by the presence of related keywords such as ``tumor" and ``cancer", leading to an overestimation of the similarity score. This example demonstrates the challenge faced by the models in accurately gauging the semantic similarity of sentences when the underlying context or focus differs, despite the presence of shared terminology.

\indent \textbf{Natural Language Inference Task.} 
In the NLI task, we focus on the MedNLI dataset and evaluate our model's performance using accuracy. On average, self-questioning prompting improves the model performance by 15.7\% and 8.8\% for zero-shot and 5-shot settings, respectively, with GPT-4 consistently outperforming Bard and GPT-3.5 across all settings.

We further investigate a pair of sentences where the gold label is ``contradiction". The first sentence states that the patient was transferred to the Neonatal Intensive Care Unit for observation, while the second sentence claims that the patient had an uneventful course. Despite the gold label, none of the models ever predict the true label, opting for ``neutral" or ``entailment" instead. The models may focus on the absence of explicit negations or conflicting keywords, leading them to overlook the more subtle contradiction. These findings highlight the need to enhance model capabilities to better understand implicit and nuanced relationships between sentences, thereby enabling more accurate predictions in complex real-world clinical scenarios.

\indent \textbf{Document Classification Task.} In the document classification task, we focus on the i2b2 2006-Smoking dataset and evaluate our model's performance using micro F1. Our analysis reveals that self-questioning prompting leads to average improvements of 1.1\% and 0.9\% for zero-shot and 5-shot settings, respectively. GPT-4 consistently delivers superior performance to Bard and GPT-3.5 in all experimental settings.

During our qualitative assessment, we investigate a patient record containing the sentence ``He is a heavy smoker and drinks 2-3 shots per day at times". All models classify the patient as a ``CURRENT SMOKER", while the patient is, in fact, a past smoker, as indicated by the subsequent descriptions of medications and the patient's improved condition. This misclassification may occur because the models focus on the explicit mention of smoking habits in the sentence, neglecting the broader context provided by the entire document. This instance highlights the need for models to take a more comprehensive approach in interpreting clinical documents by considering the overall context, rather than relying solely on individual textual cues.

\indent \textbf{Question-Answering Task.} In the QA task using the bioASQ 10b-Factoid dataset, we evaluate our model with MRR and lenient accuracy. For MRR, self-questioning prompting leads to average improvements of 4.8\% and 4.7\% for zero-shot and 5-shot settings, respectively. For lenient accuracy, the improvements are 5.8\% and 3.4\%, respectively. GPT-4 consistently outperforms Bard and GPT-3.5 across all settings.

During our qualitative exploration, we analyzed an example question: ``What is the major sequence determinant for nucleosome positioning?" The correct answer is ``G+C content"; however, the top answer from models is ``DNA sequence". This misclassification might occur because the models capture the broader context related to nucleosome positioning but fail to recognize the specific determinant, namely G+C content. The models may rely on more general associations between DNA sequences and nucleosome positioning, resulting in a less precise answer. This example underscores the necessity for models to identify fine-grained details in biomedical questions and deliver more accurate and specific responses.

\subsection{Case Study: Error Analysis}
We conduct a comprehensive error analysis on relation extraction (SemEval 2013-DDI), the most challenging task shared by all LLMs. This task is identified by calculating the median performance across all settings for a robust representation. Our process for identifying errors in relation extraction has two main stages. First, we find errors by comparing the model's outputs with the correct labels. Any differences we find are marked as errors. Next, we ask the model to explain its predictions. We manually review these explanations to spot errors, understand why they happened, and group them into specific error types. We investigate common error types and provide illustrative examples, examining the influence of prompting strategies and N-shot learning on the models' performance. This analysis highlights each model's strengths, limitations, and the role of experimental settings in improving clinical language understanding tasks.

\begin{table*}[htbp]
\centering 
  \caption{Average error type distribution for SemEval 2013-DDI across Bard, GPT-3.5, and GPT-4. Wording Ambiguity is the most common error for Bard, Lack of Context for GPT-3.5, and Negation and Qualification for GPT-4.}
  \begin{tabular}{ccccc}
    \toprule
    \multirow{2}{*}{\textbf{Error Type}} & \multirow{2}{*}{\textbf{Description}} & \multicolumn{3}{c}{\textbf{Error Proportion (\%)}}  \\
\cmidrule(lr){3-5} & & Bard & GPT-3.5 & GPT-4  \\ \midrule  
Wording Ambiguity & unclear wording & \textbf{32} & 23 & 24 \\ Lack of Context & incomplete context usage & 25 & \textbf{31} & 19 \\ Complex Interactions & multiple drug interactions & 19 & 12 & 14 \\ Negation and Qualification & \makecell{Misinterpreting \\ negation/qualification} & 8 & 27 & \textbf{25} \\ Co-reference Resolution & Misidentifying co-references & 16 & 7 & 18 \\
  \bottomrule
\end{tabular}
\label{tab: error_types}
\end{table*}

Table \ref{tab: error_types} presents the average error type distribution for the SemEval 2013-DDI task across Bard, GPT-3.5, and GPT-4. The average proportions are calculated by aggregating error frequencies for each error type across all settings and then dividing by the total number of errors for each model. The most common error type for Bard is Wording Ambiguity, accounting for 32\% of its errors, which may stem from the inherent complexity of clinical language or insufficient training data for specific drug relations. In contrast, GPT-3.5 struggles the most with Lack of Context, comprising 31\% of its errors, suggesting the model's difficulty in grasping the broader context of the input text. GPT-4's top error is Negation and Qualification, making up 25\% of its errors, possibly due to the model's limitations in understanding and processing negations and qualifications within the clinical domain. This analysis highlights the unique challenges each model faces in the relation extraction task, emphasizing the need for targeted interventions and tailored strategies to address these specific areas for improvement.

\begin{figure*}[htbp]
\centering
\includegraphics[width=\textwidth]{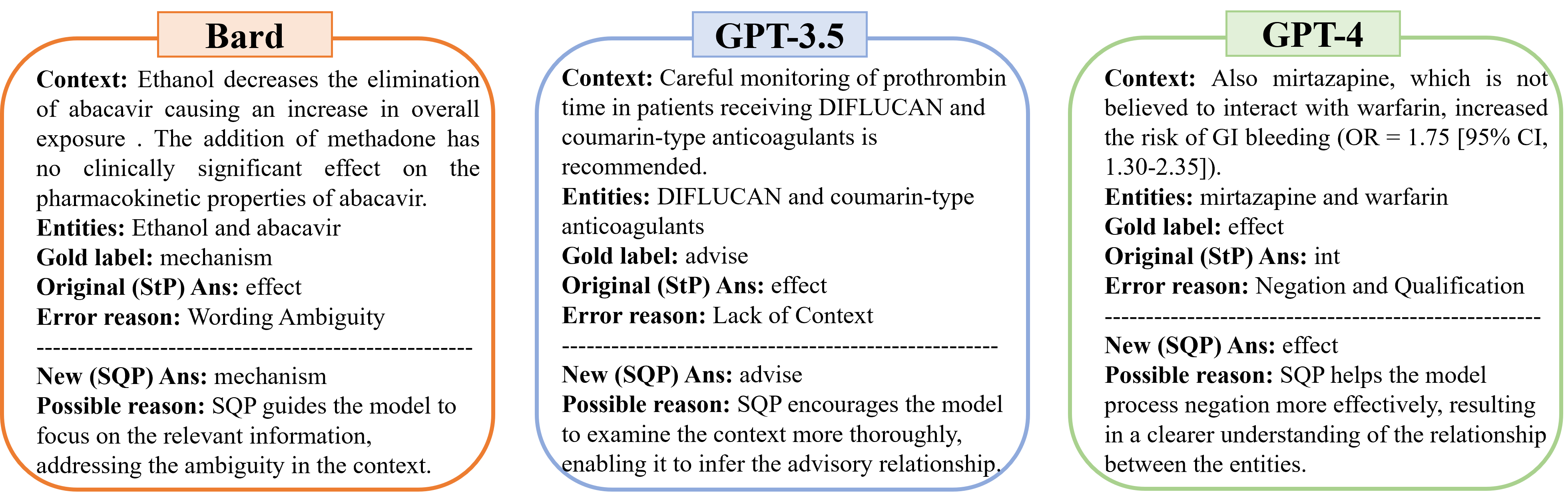}  
\caption{Error correction examples using self-questioning prompting (SQP) for Bard, GPT-3.5, and GPT-4 in the SemEval 2013-DDI dataset, compared to standard prompting (StP). Each example showcases the top error for each model and how SQP addresses these challenges. As this paper primarily focuses on the effectiveness of SQP, chain-of-thought prompting is not presented in these examples.}
\label{fig: error_example}
\end{figure*}

Specific examples presented in Figure~\ref{fig: error_example} illustrate the challenges faced by each model and how self-questioning prompting (SQP) can effectively improve their performance. SQP demonstrates its flexibility and adaptability across various model architectures by mitigating distinct error types and refining predictions. Bard sees improvements in addressing Wording Ambiguity, GPT-3.5 benefits from enhanced context utilization, and GPT-4's understanding of negation is strengthened. These examples emphasize the significance of harnessing advanced prompting techniques like SQP to bolster model performance and reveal the multifaceted challenges faced by LLMs in relation extraction tasks, particularly within the clinical domain.

Our findings highlight the potential of advanced prompting techniques, such as self-questioning prompting, in addressing model-specific errors and enhancing overall performance. These insights can be extended to various clinical language understanding tasks, guiding future research to develop more robust, accurate, and reliable models capable of processing complex clinical information and improving patient care.

\section{Discussion}
In this study, we have conducted a comprehensive evaluation of state-of-the-art large language models in the healthcare domain, including GPT-3.5, GPT-4, and Bard. We have examined the capabilities and limitations of these leading large language models across various clinical language understanding tasks such as NER, relation extraction, and QA. Our findings suggest that while LLMs have made substantial progress in understanding clinical language and achieving competitive performance across these tasks, they still exhibit notable limitations and challenges. Some of these challenges include the varying confidence levels of their responses and the difficulty in determining the trustworthiness of their generated information without human validation. Consequently, our study emphasizes the importance of using LLMs with caution as a supplement to existing workflows rather than as a replacement for human expertise. To effectively implement LLMs, clinical practitioners should employ task-specific learning strategies and prompting techniques, such as SQP, carefully designing and selecting prompts that guide the model towards better understanding and generation of relevant responses. Collaboration with experts during the development and fine-tuning of LLMs is essential to ensure accurate capture of domain-specific knowledge and sensitivity to clinical language nuances. Additionally, clinicians should be aware of the limitations and potential biases in LLMs and ensure that a human expert verifies the information they produce. By adopting a cautious approach, healthcare professionals can harness the potential of LLMs responsibly and effectively, ultimately contributing to improved patient care.

\paragraph{Limitations} While this study presents meaningful observations and sheds light on the role of large language models in the healthcare domain, there are some limitations to our work. Our study focuses on a select group of state-of-the-art LLMs, which may limit the generalizability of our findings to other models or future iterations. The performance of the proposed SQP strategy may vary depending on the tasks, prompting setup, and input-output exemplars used, suggesting that further research into alternative prompting strategies or other techniques is warranted. Our evaluation is based on a set of clinical language understanding tasks and may not cover all possible use cases in the healthcare domain, necessitating further investigation into other tasks or subdomains. Lastly, ethical and legal considerations, such as patient privacy, data security, and potential biases, are not explicitly addressed in this study. Future work should explore these aspects to ensure the responsible and effective application of LLMs in healthcare settings.

\newpage
\bibliography{sample}

\appendix
\section{Prompt Templates}~\label{prompt_template}
The templates provided herein are specific to each dataset that we utilize in our experiments. All models adhere uniformly to these templates.

\subsection{Natural Language Inference - MedNLI}
\begin{itemize}
\item Standard prompting: Assess the relationship between the following sentences: \{sentence\_1\} and \{sentence\_2\}. Is the second statement entailed by, in contradiction with, or neutral to the first statement? The answer is \{\}.
\item Chain-of-thought prompting: Identify the key ideas in \{sentence\_1\} and \{sentence\_2\}, and systematically evaluate their relationship. Categorize it as entailment if \{sentence\_2\} logically follows \{sentence\_1\}, contradiction if they oppose each other, or neutrality if unrelated. The answer is \{\}.
\item Self-questioning prompting: Given a pair of clinical sentences \{sentence\_1\} and \{sentence\_2\}, generate questions about the medical situations described. Answer these questions using basic medical knowledge and use the insights to evaluate their relationship. Categorize the relationship between \{sentence\_1\} and \{sentence\_2\} as entailment if \{sentence\_2\} logically follows \{sentence\_1\}, contradiction if they oppose each other, or neutrality if unrelated. The answer is \{\}.

\end{itemize}

\subsection{Semantic Textual Similarity - BIOSSES}
\begin{itemize}
\item Standard prompting: Assess the semantic similarity between the following two sentences: \{sentence\_1\} and \{sentence\_2\}. Provide a score on a scale from 0 (no relation) to 4 (equivalent) and can be a decimal. The similarity score is \{\}.
\item Chain-of-thought prompting: Consider clinical sentences \{sentence\_1\} and \{sentence\_2\}. Identify the key semantic relationships, and entities concepts shared by the two sentences. Then, provide a score on a scale from 0 (no relation) to 4 (equivalent), and can be a decimal. The similarity score is \{\}.
\item Self-questioning prompting: Given a pair clinical sentences \{sentence\_1\} and \{sentence\_2\}, generate a question about their medical content. Answer the question using basic medical knowledge. Keep in mind the information from the question and its answer, compare the concepts, entities, and relationships in the two sentences and rate their semantic similarity from 0 (no relation) to 4 (equivalent), and can be a decimal. The similarity score is \{\}.
\end{itemize}

\subsection{Factoid Question Answering - bioASQ 10b}
\begin{itemize}
\item Standard prompting: Answer the following factoid question and provide up to 5 candidates, ordered by decreasing confidence. Question: \{query\}. Candidate answers:
\item Chain-of-thought prompting: Given the question \{query\}, identify keywords, search for relevant information, list up to 5 candidate answers, rank them by confidence, and present an ordered list.
\item Self-questioning prompting: Given the question \{query\}, generate a related question that will help answer the question more accurately. Answer the generated question and use it to list up to 5 candidate answers that are most likely to be correct, ordered by decreasing confidence.
\end{itemize}

\subsection{Named Entity Recognition – NCBI (Disease)}
\begin{itemize}
\item Standard Prompting: Perform clinical Named Entity Recognition on the provided PubMed text \{text\} for disease name recognition and concept normalization. Your output should be in two columns, with the token in the first column and the category in the second column, separated by empty space. Tokenize each word, phrase, symbol, and punctuation as a separate token. Categorize each token as “B-Disease”, “I-Disease”, or “O”.  Note that hyphenated words or phrases should be treated as separate tokens. The order of output should follow the original text order.
\item Chain of Thought Prompting: Read and understand the following PubMed text: {text}. Identify all disease names mentioned in the text. Normalize the identified disease names to their corresponding standardized concepts. Your output should be in two columns, with the token in the first column and the category in the second column, separated by empty space. Tokenize each word, phrase, symbol, and punctuation as a separate token. Categorize each token as “B-Disease”, “I-Disease”, or “O”.  Note that hyphenated words or phrases should be treated as separate tokens. The order of output should follow the original text order.
\item Self-questioning Prompting: Given the provided PubMed text {text}, identify disease names and concepts by asking questions such as “What diseases are mentioned in the text?' and 'Which medical terms refer to diseases?” Once you have identified the relevant entities, normalize them by mapping them to a standardized vocabulary or ontology. Your output should be in two columns, with the token in the first column and the category in the second column, separated by empty space. Tokenize each word, phrase, symbol, and punctuation as a separate token. Categorize each token as “B-Disease”, “I-Disease”, or “O”.  Note that hyphenated words or phrases should be treated as separate tokens. The order of output should follow the original text order.

\end{itemize}

\subsection{Named Entity Recognition – BC5CDR (Chemical)}
\begin{itemize}
\item Standard prompting: Perform Named Entity Recognition on the provided PubMed text \{text\} for chemical entity recognition. Your output should be in two columns, with the token in the first column and the category in the second column, separated by empty space. Tokenize each word, phrase, symbol, and punctuation mark as a separate token. Categorize each token as “B-Chemical”, “I-Chemical”, or “O”. Note that hyphenated words or phrases should be treated as separate tokens. The order of output should follow the original text order.

\item Chain-of-thought prompting: Read and understand the following PubMed text: \{text\}. Identify all chemical entities. Your output should be in two columns, with the token in the first column and the category in the second column, separated by empty space. Tokenize each word, phrase, symbol, and punctuation mark as a separate token. Categorize each token as “B-Chemical”, “I-Chemical”, or “O”. Note that hyphenated words or phrases should be treated as separate tokens. The order of output should follow the original text order.
\item Self-questioning prompting: Given the provided PubMed text \{text\}, identify chemical entities by asking questions such as, “What chemicals are mentioned in the text?” Your output should be in two columns, with the token in the first column and the category in the second column, separated by empty space. Tokenize each word, phrase, symbol, and punctuation mark as a separate token. Categorize each token as “B-Chemical”, “I-Chemical”, or “O”. Note that hyphenated words or phrases should be treated as separate tokens. The order of the output should follow the original text order.

\end{itemize}

\subsection{Relation Extraction – i2b2 2010}
\begin{itemize}
\item Standard prompting: Given the context sentence {context}, identify the relationship between \{concept\_1\} and \{concept\_2\} within the sentence, and specify which category it falls under: Treatment Improves Medical Problem (TrIP), Treatment Worsens Medical Problem (TrWP), Treatment Causes Medical Problem (TrCP), Treatment is Administered for Medical Problem (TrAP), Treatment is Not Administered Because of Medical Problem (TrNAP), Test Reveals Medical Problem (TeRP), Test Conducted to Investigate Medical Problem (TeCP), or Medical Problem Indicates Medical Problem (PIP). The relationship between \{concept\_1\} and \{concept\_2\} is \{\}.
\item Chain-of-thought prompting: Given the context sentence {context}, identify the relationship between \{concept\_1\} and \{concept\_2\} within the sentence. Determine if the sentence discusses a treatment, test, or medical problem. For treatments, categorize the relationship as TrIP (improving), TrWP (worsening), TrCP (causing), TrAP (administering), or TrNAP (not administering) based on its impact on the medical problem. For tests, categorize as TeRP (revealing) or TeCP (investigating) based on the test's purpose. For medical problems, if one problem indicates another, categorize it as PIP. The relationship between \{concept\_1\} and \{concept\_2\} is \{\}.
\item Self-questioning prompting: Given the context sentence {context}, identify the relationship between \{concept\_1\} and \{concept\_2\} within the sentence. Generate questions to explore the nature of their relationship, such as whether it involves treatments improving (TrIP), worsening (TrWP), causing (TrCP), being administered for (TrAP), or not being administered due to (TrNAP) a medical problem; tests revealing (TeRP) or investigating (TeCP) a medical problem; or one medical problem indicating another (PIP). Answer the questions and use the insights to categorize the relationship between \{concept\_1\} and \{concept\_2\} as \{\}.

\end{itemize}

\subsection{Relation Extraction – DDI}
\begin{itemize}
\item Standard prompting: Given the context sentence {context}, identify the relationship between the pharmacological substances \{e\_1\} and \{e\_2\} within the sentence. Specify which category the relationship falls under: Advice, Effect, Mechanism, or Int. The relationship between \{e\_1\} and \{e\_2\} is categorized as \{\}.
\item Chain of Thought Prompting: Given the context sentence {context}, identify the relationship between the pharmacological substances \{e\_1\} and \{e\_2\} within the sentence. Analyze their interaction and classify the relationship under one of these categories: Advice, Effect, Mechanism, or Int. Consider whether the sentence advises on their use, describes their combined or counteracting effects, explains the interaction mechanism, or indicates an interaction with insufficient details. The relationship between \{e\_1\} and \{e\_2\} is categorized as \{\}.
\item Self-questioning Prompting: Given the context sentence {context}, generate questions about the relationship between the pharmacological substances \{e\_1\} and \{e\_2\} within the sentence, covering their usage advice, combined or counteracting effects, and interaction mechanisms. Answer each question and then categorize the relationship between \{e\_1\} and \{e\_2\} as Advice, Effect, Mechanism, or Int, based on the insights gained from the questions. The relationship between \{e\_1\} and \{e\_2\} is categorized as \{\}.
\end{itemize}

\subsection{Document Classification - i2b2 2006}
\begin{itemize}
\item Standard prompting: Given the patient record {record}, classify the patient’s smoking status as PAST SMOKER, CURRENT SMOKER, SMOKER, NON-SMOKER, or UNKNOWN. The patient’s smoking category is \{\}.
\item Chain-of-thought prompting: Analyze the patient record {record} to pinpoint any details about their smoking status. Reflect on the identified information, considering the patient’s smoking habits, timeframe, and available data for classification. Utilize this thought process to guide the model in determining the appropriate smoking status category among PAST SMOKER, CURRENT SMOKER, SMOKER, NON-SMOKER, or UNKNOWN. The patient’s smoking category is \{\}.
\item Self-questioning prompting: Given the patient record {record}, identify key elements in the input text related to smoking status. Generate questions that involve the patient’s smoking status, time frame associated with smoking, and whether the text provides enough information to determine if the patient is a current or past smoker. Use these questions to guide the model's response in classifying the patient’s smoking status as PAST SMOKER, CURRENT SMOKER, SMOKER, NON-SMOKER, or UNKNOWN. The patient’s smoking category is \{\}.

\end{itemize}

\end{document}